\documentclass{article}

 \usepackage[preprint]{neurips_2025}

\usepackage[utf8]{inputenc} 
\usepackage[T1]{fontenc}    
\usepackage{hyperref}       
\usepackage{url}            
\usepackage{booktabs}       
\usepackage{amsfonts}       
\usepackage{nicefrac}       
\usepackage{microtype}      
\usepackage{xcolor}         

\usepackage{amsmath} 
\usepackage{tabularx}

\usepackage{graphicx}
\usepackage{caption}
\usepackage{subcaption} 
\usepackage{float}      
\usepackage{booktabs}
\usepackage{siunitx}
\usepackage{siunitx}
\usepackage{xcolor,colortbl}

\newcommand{\score}[1]{%
  \begingroup
  \edef\val{#1}%
  \ifdim \val pt > 50pt
    \textcolor{green!40!black}{#1}%
  \else
    \textcolor{red!40!black}{#1}%
  \fi
  \endgroup
}

\usepackage[table]{xcolor} 

\title{A Coherence-Based Measure of AGI}

\author{%
  Fares Fourati \\
  KAUST\\
  \texttt{fares.fourati@kaust.edu.sa} \\
}

\begin{document}

\maketitle

\begin{abstract}
Recent approaches to evaluating \textit{Artificial General Intelligence} (AGI) typically summarize a system’s capability using the arithmetic mean of its proficiencies across multiple cognitive domains. While simple, this implicitly assumes compensability: exceptional performance in some areas can offset severe deficiencies in others. Genuine general intelligence, however, requires coherent sufficiency: balanced competence across all essential faculties.
We introduce a coherence-based measure of AGI that integrates the generalized mean over a continuum of compensability exponents. This yields an \textit{area-under-the-curve} (AUC) metric spanning arithmetic, geometric, and harmonic regimes, quantifying how robust an evaluated capability remains as compensability assumptions become stricter. Unlike the arithmetic mean, which rewards specialization, the AUC penalizes imbalance and exposes bottlenecks that constrain performance.
To illustrate the framework, we apply it to cognitive profiles derived from the Cattell–Horn–Carroll (CHC) model, showing how coherence-based aggregation highlights imbalances that are obscured by arithmetic averaging. As a second, independent example, we apply the same methodology to a set of 17 heterogeneous benchmarks, demonstrating how coherence-based evaluation can reveal unevenness even in narrower task collections. These examples show that the proposed approach offers a principled, interpretable, and stricter foundation for measuring progress toward AGI.
\end{abstract}

\section{Introduction}

\textit{Artificial General Intelligence} (AGI) is often described as the ultimate goal of Artificial Intelligence (AI) research, the creation of systems with intelligence comparable to that of humans. Yet, despite decades of progress, measuring advancement toward this goal remains elusive, largely because both \textit{intelligence} and \textit{generality} lack rigorous operational definitions \citep{chollet2019measure}. While individual capabilities can be evaluated, their integration into a single indicator of AGI is far less straightforward.

Recent work by \cite{hendrycks2025agidefinition} builds on the informal notion that “AGI is an AI that can match or exceed the cognitive versatility and proficiency of a well-educated adult,” their framework grounds evaluation in human psychometrics through the Cattell--Horn--Carroll (CHC) theory of cognitive abilities \citep{carroll1993human,mcgrew2009chc,mcgrew2023carroll}, the dominant empirical model of human intelligence. By decomposing cognition into ten broad domains (e.g., reasoning, memory, perception, and speed) and averaging performance across them, they define a unified \emph{AGI score} as the arithmetic mean of domain proficiencies.

While simple and appealing, this formulation, for example, implicitly assumes a degree of \emph{compensability}: that exceptional performance in some faculties can offset severe deficiencies in others. As a result, an AI system could appear ``general'' while failing entirely in critical domains such as reasoning or memory. This assumption conflicts with both psychometric evidence and systems theory. In human cognition, abilities are interdependent, reasoning depends on working memory, perception constrains abstraction, and learning relies on durable long-term memory~\citep{carroll1993human,mcgrew2009chc,mcgrew2023carroll}. Empirical studies in cognitive psychology consistently find that extreme imbalance across faculties is associated with functional impairment rather than high general intelligence~\citep{cattell1987intelligence}. Likewise, in complex engineered systems, overall capability is limited by the weakest component, a ``bottleneck'' or limiting-factor dynamic~\citep{keeney1993decisions,kitano2004biological}, also acknowledged by \cite{hendrycks2025agidefinition}.

We argue that general intelligence should instead reflect \emph{coherent sufficiency}: consistently high competence across all essential faculties, such that no domain falls to a low level that would preclude genuine generality. This principle resonates across disciplines, psychometrics emphasizes coherence among subtests~\cite{carroll1993human,mcgrew2009chc}, systems theory highlights limiting-factor dynamics~\cite{kitano2004biological}, and multi-criteria decision analysis formalizes non-compensatory aggregation~\cite{keeney1993decisions}. Collectively, these perspectives converge on the view that true generality arises from robustness and systemic balance rather than isolated excellence.

To formalize this intuition, we extend the arithmetic aggregation to the continuous family of \emph{generalized means} \citep{bullen2013handbook}, parameterized by a compensability exponent~$p$. When $p = 1$, the measure reduces to the arithmetic mean; as $p$ decreases, it increasingly penalizes imbalance across cognitive domains, smoothly transitioning toward non-compensatory evaluation. The resulting curve over~$p$ (e.g., Figure~\ref{fig:auc}) provides a concise diagnostic of coherence: models with flatter, higher curves exhibit more uniformly distributed competence and greater robustness to compensability assumptions.

Integrating this curve over a specified range of~$p$ yields an \emph{area-under-the-curve} (AUC) score, expressed as a percentage, that summarizes the stability of a system’s performance under increasingly strict compensability assumptions. This construction preserves continuity with prior arithmetic-mean definitions while introducing a coherence-based and theoretically grounded measure of generality.

Our contributions are threefold:
\begin{enumerate}
\item \textbf{Conceptual:} We identify \emph{compensability} as a fundamental limitation of existing AGI measurement practices and situate it within broader theories of cognitive coherence, systems interdependence, and limiting-factor dynamics. 

\item \textbf{Methodological:} We propose evaluating AGI through the full AGI\(_p\) curve, characterizing how performance behaves across compensability regimes, and through its integral, an AUC-based indicator that captures the coherence and robustness of a system’s capabilities.

\item \textbf{Illustrative:} We validate the framework in two independent settings: (i) CHC-style domain profiles for GPT-4 and GPT-5, and (ii) a heterogeneous suite of 17 benchmark scores for Gemini~3~Pro, GPT-5.1, Gemini~2.5~Pro, and Claude Sonnet~4.5. In both analyses, coherence-based aggregation exposes capability imbalances and limiting factors that arithmetic means obscure, demonstrating the practical value of the approach.
\end{enumerate}

By reframing AGI evaluation around coherence rather than compensation, this work provides a stricter, more interpretable, and more principled criterion for assessing genuine progress toward general intelligence.

\begin{figure}[t]
  \centering
  \includegraphics[width=0.55\linewidth]{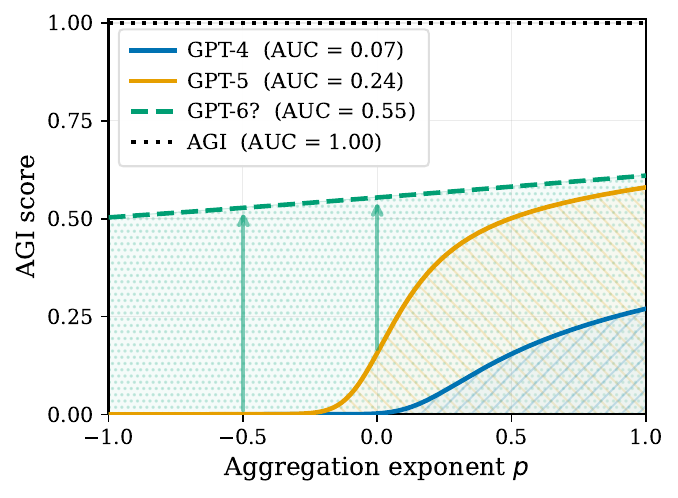}
  \caption{
    Comparison of model performance across aggregation exponents $p$.
    Curves show $\mathrm{AGI}_p$ values; shaded regions indicate the AUC.
  }
  \label{fig:auc}
\end{figure}

\section{Background}

\subsection{Psychometric Models of General Intelligence}

Human intelligence is widely understood as a structured hierarchy of interdependent abilities rather than a single undifferentiated trait.
The Cattell--Horn--Carroll (CHC) theory of cognitive abilities~\citep{carroll1993human,mcgrew2009chc,mcgrew2023carroll} provides the dominant empirical framework for modeling this structure.
CHC organizes cognition into a three-stratum hierarchy: a broad \emph{general intelligence} factor ($g$) at the apex, a set of \emph{broad abilities} (e.g., fluid reasoning, crystallized knowledge, memory, and processing speed) beneath it, and dozens of \emph{narrow abilities} at the base.
Over the past three decades, nearly all major human IQ tests have adopted the CHC approach for test construction and interpretation~\citep{keith2010cattell,mcgrew2023chc}.

The CHC framework thus operationalizes \emph{general intelligence} not as monolithic competence, but as coherent sufficiency across multiple cognitive domains.
An individual's overall intelligence score reflects consistent adequacy across reasoning, memory, perception, and other core faculties, with extreme deficiencies in any one domain typically resulting in significant functional impairment.
This interdependence of abilities motivates extending the same principle to artificial systems.

\subsection{CHC-Inspired AGI Evaluation}

Most AI benchmarks to date have assessed narrow capabilities in isolation, such as ImageNet for visual recognition~\citep{deng2009imagenet}, MATH for mathematical reasoning \citep{hendrycks2021measuring}, PIQA for commonsense inference \citep{bisk2020piqa}.
Such benchmarks measure \emph{specialization}, not \emph{generality}.
To address this, \citet{hendrycks2025agidefinition} proposed the first psychometrically grounded definition of AGI, explicitly mapping CHC's cognitive domains to artificial analogues.
Their framework decomposes machine cognition into ten broad abilities---General Knowledge, Reading and Writing, Mathematics, On-the-Spot Reasoning, Working Memory, Long-Term Memory (Storage and Retrieval), Visual Processing, Auditory Processing, and Speed---and evaluates proficiency in each through adapted psychometric tasks.

These domain scores are aggregated using the \textbf{arithmetic mean}, producing a single AGI Score between 0\% and 100\%.
For instance, \cite{hendrycks2025agidefinition} report scores of 27\% for GPT-4 \citep{achiam2023gpt} and 58\% for GPT-5, illustrating both rapid improvement and persistent gaps relative to human-level cognition.
However, this additive aggregation assumes, at least implicitly, that cognitive abilities are independent and somehow compensatory, i.e, that exceptional performance in one domain can hide complete absence in another.
As a result, the arithmetic mean can overstate overall competence, producing deceptively high composite scores for systems that appear broadly capable while still exhibiting catastrophic deficits in essential faculties such as long-term memory storage or cross-modal reasoning.

\begin{table}[h]
\centering
\caption{
Domain-level AGI scores reported by \citet{hendrycks2025agidefinition}.
K = General Knowledge; RW = Reading and Writing; M = Mathematics;
R = On-the-Spot Reasoning; WM = Working Memory; MS = Long-Term Memory Storage;
MR = Long-Term Memory Retrieval; V = Visual Processing; A = Auditory Processing;
S = Speed.
}
\begin{tabular}{
l
S[table-format=2.0]
S[table-format=2.0]
S[table-format=2.0]
S[table-format=2.0]
S[table-format=2.0]
S[table-format=2.0]
S[table-format=2.0]
S[table-format=2.0]
S[table-format=2.0]
S[table-format=2.0]
}
\toprule
\textbf{Model} & \textbf{K} & \textbf{RW} & \textbf{M} & \textbf{R} & 
\textbf{WM} & \textbf{MS} & \textbf{MR} & \textbf{V} & 
\textbf{A} & \textbf{S} \\
\midrule
GPT-4 (2023) & \score{80} & \score{60} & \score{40} & \score{0} & \score{20} & \score{0} & \score{40} & \score{0} & \score{0} & \score{30} \\
GPT-5 (2025) & \score{90} & \score{100} & \score{100} & \score{70} & \score{50} & \score{0} & \score{40} & \score{40} & \score{60} & \score{30} \\
\midrule
AGI & \score{100} & \score{100} & \score{100} & \score{100} & \score{100} & \score{100} & \score{100} & \score{100} & \score{100} & \score{100} \\
\bottomrule
\end{tabular}
\label{tab:hendrycks-results}
\end{table}

\subsection{Compensability in Complex Systems}

In multi-criteria decision theory~\citep{keeney1993decisions}, \emph{compensatory aggregation} methods assume that trade-offs between dimensions are permissible—strong performance in one criterion can offset weakness in another.
By contrast, \emph{non-compensatory aggregation} models treat dimensions as jointly constraining, such that poor performance on any critical dimension limits the overall outcome.
This framing parallels the \emph{limiting-factor principle} and analyses of robustness in complex engineered systems~\citep{kitano2004biological}, both of which emphasize that system-level functioning can be bounded by its weakest component.

In psychometrics, interdependence among cognitive abilities has been repeatedly observed.
Factor-analytic studies within the CHC framework reveal a pervasive \emph{positive manifold}—individuals who perform well in one broad ability tend to perform well across others~\citep{carroll1993human,mcgrew2009chc,mcgrew2023carroll}.
This pattern implies mutual reinforcement among abilities rather than independence.
Furthermore, empirical work shows that limitations in core processes such as working memory or processing speed can constrain higher-level reasoning and learning performance~\citep{fry2000relationships}.
Accordingly, simple averages across domains may obscure such dependencies, overstating general competence when specific foundational faculties are weak.

\subsection{Motivation for a Coherent-Based AGI Metric}

Given this background, an AGI metric should not merely \emph{sum} capabilities but ensure their coherent sufficiency, a minimal level of competence across all essential faculties.
An AI that fails completely in long-term memory or perception cannot be considered ``general,'' regardless of its average score.
In contrast, non-compensatory aggregation naturally enforces balance by penalizing uneven or ``brittle'' cognitive profiles.
The \emph{generalized mean} offers a principled mathematical family that interpolates between compensatory and non-compensatory regimes through a single parameter $p$.
This property motivates our proposed measure in the following section, which redefines general intelligence as balanced, interdependent competence across all CHC-inspired cognitive domains.

\section{A Coherence-Based Measure of General Intelligence}

Evaluating progress toward AGI requires aggregating performance
across multiple capabilities that jointly determine a system’s generality. Let
\[
s_i \in [0,100], \qquad i = 1,\dots,n,
\]
denote the normalized proficiencies of a model across $n$ benchmark dimensions. These may represent
cognitive domains, task clusters, or heterogeneous evaluations drawn from diverse modalities. Our
framework makes no assumptions about the underlying ontology: each $s_i$ need only reflect a
meaningful ability that contributes to general-purpose competence.

A widely used aggregation rule, adopted for example in the AGI definition of
\citet{hendrycks2025agidefinition}, is the arithmetic mean:
\begin{equation}
\mathrm{AGI}_{1}
=
\frac{1}{n} \sum_{i=1}^{n} s_i .
\label{eq:arithmetic}
\end{equation}
This formulation, denoted as $\mathrm{AGI}_{1}$, assumes that strengths in some faculties can compensate for weaknesses in others, a property known as \emph{compensability}. Although convenient, such additive
aggregation is misaligned with both psychometric evidence and limiting-factor dynamics in complex
systems. Severe deficiencies in essential faculties, e.g., reasoning, memory, or perception, often
impose system-level bottlenecks that cannot be compensated by strengths elsewhere, causing
$\mathrm{AGI}_1$ to substantially overestimate holistic capability.

\subsection{Generalized Means and Compensability}

To relax this compensatory assumption, we generalize Eq.~\eqref{eq:arithmetic} to generalized-mean (power-mean)
family \citep{bullen2013handbook}, parameterized by a compensability exponent $p \in \mathbb{R}$, where each value of $p$ corresponds to a distinct \emph{coherence regime} for general intelligence, controlling the degree to which strengths can offset weaknesses:
\begin{equation}
\mathrm{AGI}_p =
\begin{cases}
\left( \dfrac{1}{n} \sum_{i=1}^{n} \max\!\left(s_i, \varepsilon\right)^{p} \right)^{\!\!1/p}, & p \neq 0, \\[10pt]
\left( \prod_{i=1}^{n} \max\!\left(s_i, \varepsilon\right) \right)^{\!1/n}, & p = 0 ,
\end{cases}
\label{eq:agip}
\end{equation}

where $\varepsilon>0$ is a small stability constant (set to $10^{-6}$ in all experiments) that prevents
numerical collapse when any $s_i = 0$ and can be viewed as a minimal detectable competence or a noise floor in domain measurement.  

The exponent $p$ determines the allowed degree of compensability:
\begin{itemize}
    \item $p = 1$: arithmetic mean (strongly compensatory),
    \item $p = 0$: geometric mean (moderately non-compensatory),
    \item $p = -1$: harmonic mean (strongly non-compensatory),
    \item $p \to -\infty$: minimum operator (strict bottleneck).
\end{itemize}

As $p$ decreases, $\mathrm{AGI}_p$ shifts from measuring \emph{average proficiency} to measuring \emph{coherent sufficiency}, requiring balanced competence across all essential dimensions. This reflects the intuition that general intelligence is not characterized by isolated peaks but by the absence of critical weaknesses.

The limiting case $p \to -\infty$ yields a fully non-compensatory measure equal to the weakest domain score, but is often too sparse to be practical: the overall score changes only when the minimum dimension improves. Even the harmonic mean ($p=-1$) can collapse toward zero on uneven profiles. Conversely, permissive regimes such as $p=1$ can mask catastrophic deficits, yielding inflated scores despite near-zero performance in key areas.

These considerations motivate restricting attention to a moderate range, such as $p \in [-1,1]$, which enforces meaningful non-compensatory behavior while preserving sensitivity and avoiding degeneracy. Within this interval, the $\mathrm{AGI}_p$ curve provides a concise diagnostic of coherence: models with flatter, higher curves exhibit more uniformly distributed competence and greater robustness to compensability assumptions.

For a single scalar summary of progress toward AGI, one may integrate $\mathrm{AGI}_p$ over the chosen $p$-range, yielding a measure that captures both robustness and coherence.

\subsection{An Integrated Measure of Coherence: \texorpdfstring{$\mathrm{AGI}_{\mathrm{AUC}}$}{AGI AUC}}

To summarize robustness across compensability regimes, we define the $\mathrm{AGI}_{\mathrm{AUC}}$ metric:
\begin{equation}
\mathrm{AGI}_{\mathrm{AUC}}
=
\frac{1}{p_{\max}-p_{\min}}
\int_{p_{\min}}^{p_{\max}}
\mathrm{AGI}_p \, dp .
\label{eq:agiauc}
\end{equation}
Throughout this work, we set $p_{\min}=-1$ and $p_{\max}=1$, spanning the range from strongly non-compensatory to strongly compensatory aggregation. The integral can be evaluated numerically using the composite trapezoidal rule over a uniform grid in $p$.

The $\mathrm{AGI}_p$ family highlights the characteristic ``jaggedness'' of modern AI systems: high proficiency in some domains coexists with catastrophic failures in others. 

$\mathrm{AGI}_{\mathrm{AUC}}$ measures \emph{coherence} and \emph{stability across compensability regimes}. Systems with high $\mathrm{AGI}_1$ but low $\mathrm{AGI}_{\mathrm{AUC}}$ rely on narrow excellence and lack balanced, integrated capability. In contrast, a system approaching general intelligence would display a uniformly high $\mathrm{AGI}_p$ curve across the entire range of $p$, indicating robustness even under stricter non-compensatory evaluation. Thus, $\mathrm{AGI}_{\mathrm{AUC}}$ provides a principled and coherence-sensitive summary of general intelligence, capturing both the breadth and the structural balance of a system's abilities.

\section{Results on the CHC-Based Domain Scores}
\label{results_chc}

\paragraph{Setup.}
We apply the proposed framework to the published AGI domain scores of \citet{hendrycks2025agidefinition} for GPT-4~\citep{achiam2023gpt} and GPT-5~(2025).  
For completeness, these baseline scores are reproduced in Table~\ref{tab:hendrycks-results}.  
The floor constant was fixed at $\varepsilon = 10^{-6}$, and the compensability interval was set to $p \in [-1,1]$.  
We report $\mathrm{AGI}_{-1}$, $\mathrm{AGI}_{-0.5}$, $\mathrm{AGI}_0$, $\mathrm{AGI}_{0.5}$, and $\mathrm{AGI}_1$, along with their corresponding $\mathrm{AGI}_p$ curves and the aggregate $\mathrm{AGI}_{\mathrm{AUC}}$ measure. To assess external alignment, we also compare the resulting $\mathrm{AGI}_{\mathrm{AUC}}$ values with performance on challenging intelligence benchmarks.

The analysis in this section builds directly on the ten normalized domain scores reported in \citet{hendrycks2025agidefinition}.  
In Appendix~\ref{appendix:norm}, we document the origin of these values and compute several alternative aggregates, including weighted and unweighted geometric means.  
We keep this extended discussion in the appendix to maintain focus on the primary contribution of this paper: demonstrating how coherence-based aggregation provides a more principled framework for assessing AGI.

\paragraph{Results.}

\begin{table}[t]
  \centering
  \caption{
    Key scores per model (in \%). 
    $\mathrm{AGI}_1$ corresponds to the arithmetic mean ($p{=}1$), 
    $\mathrm{AGI}_{0.5}$ to the quasi-arithmetic mean ($p{=}0.5$), 
    $\mathrm{AGI}_0$ to the geometric mean ($p{=}0$), 
    $\mathrm{AGI}_{-0.5}$ to the inverse-quasi mean ($p{=}-0.5$), 
    $\mathrm{AGI}_{-1}$ to the harmonic mean ($p{=}-1$), and 
    $\mathrm{AGI}_{\text{AUC}}$ to the area under the $\mathrm{AGI}_p$ curve. 
    Values are expressed as percentages relative to the ideal AGI reference model ($=100\%$).
  }
  \label{tab:key-scores-extended}
  \begin{tabular}{lccccc|c}
    \toprule
    \textbf{Model} 
      & {\textbf{AGI}$_{\mathbf{1}}$} 
      & {\textbf{AGI}$_{\mathbf{0.5}}$ } 
      & {\textbf{AGI}$_{\mathbf{0}}$} 
      & {\textbf{AGI}$_{\mathbf{-0.5}}$ } 
      & {\textbf{AGI}$_{\mathbf{-1}}$ } 
      & {\textbf{AGI}$_{\text{AUC}}$ (ours) } \\
    \midrule
    GPT-4 (2023)   & \score{27} & \score{16} & \score{0} & \score{0} & \score{0} & \textbf{\score{7}} \\
    GPT-5 (2025)  & \score{58} & \score{50} & \score{16} & \score{0} & \score{0} & \textbf{\score{24}} \\
    \midrule
    AGI     & \score{100} & \score{100} & \score{100} & \score{100} & \score{100} & \score{100} \\
    \bottomrule
  \end{tabular}
\end{table}

Figure~\ref{fig:auc} and Table~\ref{tab:key-scores-extended} summarize how model performance varies under different compensability assumptions.  
When $p = 1$, the arithmetic mean ($\mathrm{AGI}_1$) reproduces the values reported in \citet{hendrycks2025agidefinition}.  
Under this compensatory definition, GPT-5 appears to have made substantial progress over GPT-4 (\score{58}\% vs.~\score{27}\%), suggesting meaningful gains in overall capability and surpassing the halfway mark toward human-level proficiency.

However, as compensability decreases, this apparent advantage diminishes sharply.  
The geometric mean ($\mathrm{AGI}_0$) collapses to near zero for both GPT-4 and GPT-5, reflecting the persistence of minimal proficiency in several cognitive domains.  
This pattern indicates that both systems rely heavily on a subset of strong faculties rather than demonstrating balanced competence across domains.  
Consequently, the arithmetic mean tends to overstate progress by averaging across highly uneven abilities, producing an inflated signal of movement toward AGI.

The area-under-the-curve measure, $\mathrm{AGI}_{\mathrm{AUC}}$, integrates performance across $p \in [-1,1]$ to capture overall coherence.  
GPT-5 achieves a roughly threefold improvement over GPT-4 (\score{24}\% vs.~\score{7}\%), reflecting greater but still limited robustness.  
The $\mathrm{AGI}_p$ curves in Figure~\ref{fig:auc} exhibit steep declines for negative $p$ values, indicating that weaknesses in domains such as long-term memory and adaptive reasoning continue to act as system-wide bottlenecks.  
A system approaching genuine generality would instead maintain a flatter and higher $\mathrm{AGI}_p$ curve across the full range of $p$, indicating resilience under increasingly strict coupling regimes.  
In the speculative case of ``GPT-6?'', shown in In Figure~\ref{fig:auc}, we simulate the effect of mitigating GPT-5’s principal bottleneck in long-term memory storage (MS; see Table~\ref{tab:hendrycks-results}) by increasing its domain score from \score{0}\% to \score{30}\%, while keeping all other domain scores unchanged. 
Both $\mathrm{AGI}_p$ and $\mathrm{AGI}_{\mathrm{AUC}}$ increase substantially, suggesting that even modest improvements in the weakest faculties can yield disproportionate gains in overall coherence as captured by the proposed metrics.  

These results motivate the use of $\mathrm{AGI}_{\mathrm{AUC}}$ as a more reliable summary metric.  
Strict aggregations such as $\mathrm{AGI}_{-1}$ or $\mathrm{AGI}_{-0.5}$ are excessively punitive, collapsing uneven but meaningful progress to near-zero values, whereas optimistic measures such as $\mathrm{AGI}_1$ or $\mathrm{AGI}_{0.5}$ exaggerate progress by ignoring structural dependencies.  
By integrating across compensability regimes, $\mathrm{AGI}_{\mathrm{AUC}}$ jointly captures both \emph{breadth} and \emph{balance}, yielding a stable and interpretable indicator of holistic capability.  
It neither collapses for incomplete systems nor inflates scores for narrow specialists, making it a robust and coherence-sensitive measure of emerging general intelligence.

\paragraph{Alignment with external benchmarks.}
We compare the $\mathrm{AGI}_p$ metrics with challenging external reasoning benchmarks.  
The ARC-AGI-2 benchmark~\citep{chollet2019measure,chollet2025arc}, which emphasizes out-of-distribution reasoning and cross-domain abstraction, reveals a clear calibration effect.  
While GPT-5 achieves an arithmetic-mean score of $\mathrm{AGI}_1 = \score{58}\%$, suggesting near-human generality, its coherence-aware score of $\mathrm{AGI}_{\mathrm{AUC}} = \score{24}\%$ aligns much more closely with ARC-AGI-2 results, where GPT-5~(Pro) reaches approximately \score{18}\% and GPT-5~(High) about \score{10}\%.  
Similarly, the BIG-Bench Extra Hard evaluations~\citep{kazemi2025big}, which probe broader reasoning proficiency across linguistically diverse tasks, report a GPT-4 score of roughly \score{6}\%, closely matching its $\mathrm{AGI}_{\mathrm{AUC}} = \score{7}\%$, unlike the arithmetic mean of \score{27}\%.

This correspondence supports the interpretation that $\mathrm{AGI}_{\mathrm{AUC}}$ more faithfully captures \emph{functional coherence}, the capacity to sustain competence across diverse cognitive domains, whereas the arithmetic mean $\mathrm{AGI}_1$ tends to overstate progress by rewarding isolated peaks of specialization. Accordingly, $\mathrm{AGI}_{\mathrm{AUC}}$ serves as a more conservative indicator of emerging generality, aligning closely with independent measures of out-of-distribution reasoning such as ARC-AGI-2 and BIG-Bench Hard.


\section{Beyond CHC-Domain Evaluation}
\label{beyond_chc_summary}

While the previous section focuses on the ten CHC-inspired cognitive domains defined by \citet{hendrycks2025agidefinition}, the coherence-based framework developed in this work is not tied to that taxonomy. As demonstrated in Appendix~\ref{appendix_beyond}, we further evaluate the same generalized-mean family $\mathrm{AGI}_p$ and the integrated coherence metric $\mathrm{AGI}_{\mathrm{AUC}}$ on a heterogeneous set of 17 benchmarks drawn from the publicly released \emph{Gemini~3~Pro Model Evaluation Report}~\citep{deepmind2025gemini3pro}, for GPT-5.1, Gemini 3 Pro, Claude Sonnet 2.5 Pro, and Claude Sonnet 4.5 \citep{claude_sonnet4_5}. These benchmarks span symbolic reasoning, scientific problem solving, mathematical tasks, coding and agentic behavior, multimodal abstraction, and long-context understanding, illustrating the flexibility of the framework across evaluation settings.

The extended analysis reveals patterns that closely mirror those observed in the CHC-domain setting. Aggregation with $p = 1$ (i.e., the arithmetic mean) again yields optimistic assessments of overall capability by masking imbalances across tasks, whereas negative-$p$ aggregation exposes persistent bottlenecks. Models with similar average performance diverge sharply in coherence: their $\mathrm{AGI}_p$ curves show steep declines for $p < 0$, and $\mathrm{AGI}_{\mathrm{AUC}}$ provides a clearer and more stable ordering of holistic competence. These results reinforce that coherence captures whether a system’s abilities form a balanced, general-purpose profile rather than a collection of narrow skill peaks.

At the same time, while coherence-based aggregation provides a substantially more principled and informative summary than the arithmetic mean, any benchmark-derived score must ultimately be interpreted in the context of the benchmark suite it aggregates. To meaningfully quantify “progress toward AGI,’’ the benchmarks themselves must span the full spectrum of essential cognitive faculties. Expanding benchmark coverage—across continual learning, long-horizon reasoning, memory formation, planning, and other fundamental abilities, remains an important direction for the community. Within any given benchmark set, however, our coherence-based aggregation offers a robust and conservative measure of generality: it penalizes uneven skill profiles, highlights hidden bottlenecks, and rewards the balance and robustness that are core to general intelligence.

\section{Discussion}

The results indicate that the arithmetic mean used in prior definitions of AGI (\(\mathrm{AGI}_1\)) 
is a fragile indicator of generality. It rewards specialization and conceals structural weaknesses. Under coherence-aware formulations 
(\(\mathrm{AGI}_0\), \(\mathrm{AGI}_{-1}\), and \(\mathrm{AGI}_{\text{AUC}}\)), 
the same systems score far lower, revealing that contemporary models remain unevenly developed across 
core cognitive domains and skills.

\paragraph{The current CHC-derived score is stricter yet still optimistic.}
The coherence-based framework reveals that current estimates of general intelligence in frontier models are likely inflated by additive aggregation.  
While GPT-5 achieves an $\mathrm{AGI}_{1}$ (arithmetic mean) of $\score{58}\%$, suggesting near-human breadth, 
its coherence-adjusted $\mathrm{AGI}_{\mathrm{AUC}}$ falls to only $\score{24}\%$.  
Even this value should be regarded as an \emph{upper bound} on general competence.  
As shown in Appendix~\ref{appendix:norm}, the ten constituent cognitive scores from \citet{hendrycks2025agidefinition} 
themselves display substantial inflation when aggregated arithmetically.  
When recalculated using geometric or unweighted means, several domains collapse: 
auditory processing drops from roughly $\score{60}\%$ to near \score{0}\%, visual processing from $\score{40}\%$ to near $\score{0}\%$, 
processing speed from $\score{30}\%$ to near $\score{0}\%$, and on-the-spot reasoning from $\score{70}\%$ (arithmetic) to 
$\score{19}\%$ (weighted geometric) or $\score{5}\%$ (unweighted).  
These discrepancies expose a systematic overestimation of generality due to compensatory aggregation, 
where strong subskills mask critical deficiencies elsewhere.  
Accordingly, the nominal $\mathrm{AGI}_{\mathrm{AUC}}$ of $\score{24}\%$ should be interpreted as conservative 
yet still optimistic, given (potential) inflation at the domain-score level. 

\paragraph{Rethinking progress toward AGI.}
The divergence between \(\mathrm{AGI}_1\) and coherence-based metrics highlights a key distinction between 
\emph{breadth of performance} and \emph{integration of capability}.  
Improvements in select domains inflate arithmetic averages but contribute little to genuine generality 
if other faculties remain weak. This mirrors refinements in human psychometrics, 
where single-factor IQ models evolved into hierarchical frameworks 
\citep{carroll1993human,mcgrew2009chc,mcgrew2023carroll} 
to capture both domain-specific and global ability.  
AI evaluation requires a similar shift, from measuring \emph{average proficiency} 
to assessing \emph{functional sufficiency} across interdependent cognitive dimensions.

\paragraph{Toward coherence-based evaluation.}
The generalized-mean formulation offers a continuous lens for exploring compensability assumptions.  
Across this spectrum, \(\mathrm{AGI}_{\text{AUC}}\) emerges as the most stable and informative single measure: 
it integrates performance across optimistic and strict regimes, balancing sensitivity to weaknesses 
with tolerance for partial unevenness.  
Systems that achieve high \(\mathrm{AGI}_1\) but low \(\mathrm{AGI}_0\) or \(\mathrm{AGI}_{\text{AUC}}\) 
should thus be interpreted as \emph{specialized} rather than coherent generalists.  
Conversely, genuine progress toward AGI would manifest as a flatter, consistently high \(\mathrm{AGI}_p\) curve, 
indicating resilience under increasingly strict coupling of abilities.  
Future benchmarks and model cards should therefore report both 
\(\mathrm{AGI}_1\) (average proficiency) and coherence-oriented measures 
such as \(\mathrm{AGI}_{\text{AUC}}\) 
to disentangle breadth from balance.

\paragraph{On metric pluralism.}
Although coherence-based measures expose the structural limitations of arithmetic averaging, no single scalar, including $\mathrm{AGI}_{\mathrm{AUC}}$, can fully capture general intelligence. Each aggregate reflects a complementary facet of capability: the arithmetic mean highlights breadth, negative-$p$ regimes reveal bottlenecks, and the full $\mathrm{AGI}_p$ curve provides a visual diagnostic of how performance behaves under varying compensability assumptions. Rather than serving as a single leaderboard number, the continuum over $p$ forms a multidimensional assessment toolkit. External benchmarks such as ARC-AGI-2~\citep{chollet2025arc} and BIG-Bench Extra Hard~\citep{kazemi2025big} further complement this analysis, and can themselves be integrated using the same generalized-mean framework to yield coherence-aware summaries of performance.

\paragraph{Interpretive implications.}
Low curves or AUC-based scores do not imply an absence of generalization, 
but rather highlight fragility when tasks demand multi-domain coordination.  
The coherence perspective thus bridges psychometric and computational traditions, 
emphasizing integration and stability over isolated domain success.
The goal of proposing a stricter, coherence-based measure is not to diminish 
the remarkable capabilities of current GPT-class systems, but to sharpen the lens 
through which progress toward AGI is assessed.  
True general intelligence should demonstrate balanced competence in reasoning, abstraction, generalization, planning, and memory, the capacities that underpin adaptability and understanding beyond pattern replication.  
By tightening the definition of generality, this framework aims to steer progress toward integrated, cross-domain coherence, an intelligence that not only performs well, but \emph{reasons, learns, and plans} as a unified system.


\paragraph{Adaptive evaluation framework.}
A central advantage of the proposed coherence-based measure is its generality: it applies uniformly across heterogeneous benchmark families without committing to a fixed cognitive taxonomy. Our formulation allows for any collection of tasks, whether CHC-aligned domains (as in Section~\ref{results_chc}), symbolic reasoning suites, multimodal assessments, or mixed benchmark portfolios (as in Section~\ref{beyond_chc_summary}), to be aggregated using the same generalized-mean family $\mathrm{AGI}_p$. The metric depends only on normalized task scores, not on assumptions about human-inspired structure or domain boundaries. As demonstrated in Appendix~\ref{appendix_beyond}, this enables seamless integration of diverse evaluations (e.g., Humanity's Last Exam \citep{phan2025humanity} and ARC-AGI-2 \citep{chollet2025arc}), producing coherence-based summaries that remain practical even as benchmarks evolve. By grounding assessment in task-level constructs while maintaining a taxonomy-agnostic aggregation rule, the framework is inherently adaptable: new benchmarks, modalities, or task definitions can be incorporated without modifying the underlying notion of general intelligence.



\section{Conclusion}

The arithmetic-mean definition of general intelligence, while simple and interpretable, conceals the structural imbalance of current AI systems. By generalizing this formulation to a continuum of compensability exponents, we reveal that apparent progress toward AGI is often confined to narrow strengths rather than coherent, system-wide competence. The proposed family of $\mathrm{AGI}_p$ measures, and in particular the integrated $\mathrm{AGI}_{\text{AUC}}$, provides a stricter and more interpretable framework for assessing generality. 

Our results show that coherence, the balanced sufficiency of abilities across domains, is a more faithful indicator of general intelligence than arithmetic breadth alone. Genuine progress toward AGI will thus manifest not as higher averages, but as flatter and elevated $\mathrm{AGI}_p$ curves reflecting resilient, interdependent capabilities. Incorporating coherence-based metrics such as $\mathrm{AGI}_{\text{AUC}}$ into future benchmarks can align empirical evaluation with the intended meaning of ``general'' in AGI.

\newpage
\bibliographystyle{unsrtnat}
\bibliography{biblio}



\newpage
\appendix

\providecommand{\score}[1]{#1}

\section{Normalized Cognitive Capacity Tables (Percentage Form)}
\label{appendix:norm}

This appendix provides a detailed quantitative analysis of model performance across the ten cognitive domains defined in the \textit{A Definition of AGI} framework \citep{hendrycks2025agidefinition}.
All subdomain values are drawn directly from \citet{hendrycks2025agidefinition} and are normalized to express the percentage of human-level proficiency attained within each subdomain.
Subdomain weights follow the fixed allocations specified in that framework, reflecting each domain’s theoretical contribution to overall general intelligence; these weights may be refined in future extensions of the methodology.


Each subdomain carries a maximum weight that reflects its intended influence on the composite score. 
Subdomain raw results (drawb directly from \citep{hendrycks2025agidefinition}) are normalized as:
\[
\text{Normalized Score (\%)} = \frac{\text{Raw Score}}{\text{Weight}} \times 100,
\]
so that \(100\%\) corresponds to human-equivalent proficiency within that subdomain.

\paragraph{Four domain-level aggregates.}
To represent both breadth and robustness, we report:
\begin{itemize}
    \item \textbf{AM} — Unweighted arithmetic mean (breadth of capability).
    \item \textbf{WAM} — Weighted arithmetic mean using subdomain weight shares.
    \item \textbf{GM} — Unweighted geometric mean (robustness without weighting).
    \item \textbf{WGM} — Weighted geometric mean using subdomain weights.
\end{itemize}

For geometric means, let \(p_i = \text{score}_i / 100 \in [0,1]\), \(\tilde{w}_i = w_i/\sum_j w_j\), and use a stabilizer \(\varepsilon = 10^{-6}\) to avoid collapse when any \(p_i=0\):
\[
\mathrm{GM} = 100 \times \Big(\prod_i \max(p_i,\varepsilon)\Big)^{1/n}, \qquad
\mathrm{WGM} = 100 \times \prod_i \max(p_i,\varepsilon)^{\tilde{w}_i}.
\]

\subsection{General Knowledge (K)}

Table~\ref{tab:K_percent} shows high crystallized knowledge for both models, with GPT-5 improving cultural coverage to \(\score{50}\) and achieving strong robustness (GM/WGM \(\score{87.1}\)). GPT-4’s zero in culture drops geometric aggregates, exposing imbalance despite an \(\score{80.0}\) AM.

\begin{table}[H]
\small
\centering
\caption{General Knowledge (K): Normalized Subdomains with Weights and Aggregates}
\label{tab:K_percent}
\begin{tabular}{lccccc|cccc}
\toprule
\textbf{Model} & Common & Science & Soc.\ Sci. & History & Culture & \textbf{AM} & \textbf{WAM} & \textbf{GM} & \textbf{WGM} \\
\midrule
\textbf{Weight} & 20 & 20 & 20 & 20 & 20 & -- & -- & -- & -- \\
\midrule
GPT-4 & \score{100} & \score{100} & \score{100} & \score{100} & \score{0} & \score{80.0} & \score{80.0} & \score{6.3} & \score{6.3} \\
GPT-5 & \score{100} & \score{100} & \score{100} & \score{100} & \score{50} & \score{90.0} & \score{90.0} & \score{87.1} & \score{87.1} \\
\bottomrule
\end{tabular}
\end{table}


\subsection{Reading and Writing (RW)}

As detailed in Table~\ref{tab:RW_percent}, GPT-4’s uneven orthography/usage yields a brittle GM (\(\score{2.2}\)). GPT-5 is uniformly strong across all subskills, reflected in perfect aggregates.

\begin{table}[H]
\small
\centering
\caption{Reading and Writing (RW): Normalized Subdomains with Weights and Aggregates}
\label{tab:RW_percent}
\begin{tabular}{lcccc|cccc}
\toprule
\textbf{Model} & Letters & Reading & Writing & Usage & \textbf{AM} & \textbf{WAM} & \textbf{GM} & \textbf{WGM} \\
\midrule
\textbf{Weight} & 10 & 30 & 30 & 30 & -- & -- & -- & -- \\
\midrule
GPT-4 & \score{0} & \score{67} & \score{100} & \score{33} & \score{50.0} & \score{60.0} & \score{2.2} & \score{16.0} \\
GPT-5 & \score{100} & \score{100} & \score{100} & \score{100} & \score{100.0} & \score{100.0} & \score{100.0} & \score{100.0} \\
\bottomrule
\end{tabular}
\end{table}

\subsection{Mathematical Ability (M)}

Table~\ref{tab:M_percent} indicates GPT-4’s failures in geometry/calculus drive near-zero GM/WGM, while GPT-5 closes all gaps, consistent with stronger symbolic and quantitative abstraction.

\begin{table}[H]
\small
\centering
\caption{Mathematical Ability (M): Normalized Subdomains with Weights and Aggregates}
\label{tab:M_percent}
\begin{tabular}{lccccc|cccc}
\toprule
\textbf{Model} & Arithmetic & Algebra & Geometry & Prob. & Calculus & \textbf{AM} & \textbf{WAM} & \textbf{GM} & \textbf{WGM} \\
\midrule
\textbf{Weight} & 20 & 20 & 20 & 20 & 20 & -- & -- & -- & -- \\
\midrule
GPT-4 & \score{100} & \score{50} & \score{0} & \score{50} & \score{0} & \score{40.0} & \score{40.0} & \score{0.3} & \score{0.3} \\
GPT-5 & \score{100} & \score{100} & \score{100} & \score{100} & \score{100} & \score{100.0} & \score{100.0} & \score{100.0} & \score{100.0} \\
\bottomrule
\end{tabular}
\end{table}

\subsection{On-the-Spot Reasoning (R)}

In Table~\ref{tab:R_percent}, GPT-5 improves in deduction, ToM, and planning, but adaptive reasoning remains at \(\score{0}\). Low GM/WGM (\(\score{5.5}/\score{19.0}\)) indicate fragile, non-transferable reasoning.

\begin{table}[H]
\small
\centering
\caption{On-the-Spot Reasoning (R): Normalized Subdomains with Weights and Aggregates}
\label{tab:R_percent}
\begin{tabular}{lccccc|cccc}
\toprule
\textbf{Model} & Deduction & Induction & ToM & Planning & Adapt. & \textbf{AM} & \textbf{WAM} & \textbf{GM} & \textbf{WGM} \\
\midrule
\textbf{Weight} & 20 & 40 & 20 & 10 & 10 & -- & -- & -- & -- \\
\midrule
GPT-4 & \score{0} & \score{0} & \score{0} & \score{0} & \score{0} & \score{0.0} & \score{0.0} & \score{0.0} & \score{0.0} \\
GPT-5 & \score{100} & \score{50} & \score{100} & \score{100} & \score{0} & \score{70.0} & \score{70.0} & \score{5.5} & \score{19.0} \\
\bottomrule
\end{tabular}
\end{table}

\subsection{Working Memory (WM)}

As seen in Table~\ref{tab:WM_percent}, capacity is dominated by text; GPT-5 shows partial progress in visual and cross-modal maintenance, but GM/WGM remain low, signaling a bottleneck.

\begin{table}[H]
\small
\centering
\caption{Working Memory (WM): Normalized Subdomains with Weights and Aggregates}
\label{tab:WM_percent}
\begin{tabular}{lcccc|cccc}
\toprule
\textbf{Model} & Textual & Auditory & Visual & Cross-Mod. & \textbf{AM} & \textbf{WAM} & \textbf{GM} & \textbf{WGM} \\
\midrule
\textbf{Weight} & 20 & 20 & 40 & 20 & -- & -- & -- & -- \\
\midrule
GPT-4 & \score{100} & \score{0} & \score{0} & \score{0} & \score{25.0} & \score{20.0} & \score{0.0} & \score{0.0} \\
GPT-5 & \score{100} & \score{0} & \score{25} & \score{50} & \score{43.8} & \score{40.0} & \score{1.9} & \score{3.2} \\
\bottomrule
\end{tabular}
\end{table}


\subsection{Long-Term Memory Storage (MS)}

Table~\ref{tab:MS_percent} confirms no durable storage: all subdomains are \(\score{0}\) for both models. This undermines coherence and constrains non-compensatory aggregates.

\begin{table}[H]
\small
\centering
\caption{Long-Term Memory Storage (MS): Normalized Subdomains with Weights and Aggregates}
\label{tab:MS_percent}
\begin{tabular}{lccc|cccc}
\toprule
\textbf{Model} & Assoc. & Meaningful & Verbatim & \textbf{AM} & \textbf{WAM} & \textbf{GM} & \textbf{WGM} \\
\midrule
\textbf{Weight} & 40 & 30 & 30 & -- & -- & -- & -- \\
\midrule
GPT-4 & \score{0} & \score{0} & \score{0} & \score{0.0} & \score{0.0} & \score{0.0} & \score{0.0} \\
GPT-5 & \score{0} & \score{0} & \score{0} & \score{0.0} & \score{0.0} & \score{0.0} & \score{0.0} \\
\bottomrule
\end{tabular}
\end{table}

\subsection{Long-Term Memory Retrieval (MR)}

As shown in Table~\ref{tab:MR_percent}, partial fluency is offset by hallucination-prone retrieval, leaving GM/WGM near zero. Retrieval weaknesses reinforce the storage gap in Table~\ref{tab:MS_percent}.

\begin{table}[H]
\small
\centering
\caption{Long-Term Memory Retrieval (MR): Normalized Subdomains with Weights and Aggregates}
\label{tab:MR_percent}
\begin{tabular}{lcc|cccc}
\toprule
\textbf{Model} & Fluency & Halluc.\ Avoid. & \textbf{AM} & \textbf{WAM} & \textbf{GM} & \textbf{WGM} \\
\midrule
\textbf{Weight} & 60 & 40 & -- & -- & -- & -- \\
\midrule
GPT-4 & \score{67} & \score{0} & \score{33.5} & \score{40.2} & \score{0.1} & \score{0.3} \\
GPT-5 & \score{67} & \score{0} & \score{33.5} & \score{40.2} & \score{0.1} & \score{0.3} \\
\bottomrule
\end{tabular}
\end{table}

\subsection{Visual Processing (V)}

Table~\ref{tab:V_percent} indicates progress for GPT-5 in perception/generalization, but reasoning/spatial remain at \(\score{0}\). Low GM/WGM reflect this limitation.

\begin{table}[H]
\small
\centering
\caption{Visual Processing (V): Normalized Subdomains with Weights and Aggregates}
\label{tab:V_percent}
\begin{tabular}{lcccc|cccc}
\toprule
\textbf{Model} & Percep. & Gen. & Reason. & Spatial & \textbf{AM} & \textbf{WAM} & \textbf{GM} & \textbf{WGM} \\
\midrule
\textbf{Weight} & 40 & 30 & 20 & 10 & -- & -- & -- & -- \\
\midrule
GPT-4 & \score{0} & \score{0} & \score{0} & \score{0} & \score{0.0} & \score{0.0} & \score{0.0} & \score{0.0} \\
GPT-5 & \score{50} & \score{67} & \score{0} & \score{0} & \score{29.3} & \score{40.1} & \score{0.1} & \score{1.1} \\
\bottomrule
\end{tabular}
\end{table}

\subsection{Auditory Processing (A)}

In Table~\ref{tab:A_percent}, GPT-5 reaches strong speech recognition and partial voice discrimination, but lacks phonetic/rhythmic/musical competence. The GM/WGM confirms incomplete auditory integration.

\begin{table}[H]
\small
\centering
\caption{Auditory Processing (A): Normalized Subdomains with Weights and Aggregates}
\label{tab:A_percent}
\begin{tabular}{lccccc|cccc}
\toprule
\textbf{Model} & Phonetic & Speech Rec. & Voice & Rhyth. & Musical & \textbf{AM} & \textbf{WAM} & \textbf{GM} & \textbf{WGM} \\
\midrule
\textbf{Weight} & 10 & 40 & 30 & 10 & 10 & -- & -- & -- & -- \\
\midrule
GPT-4 & \score{0} & \score{0} & \score{0} & \score{0} & \score{0} & \score{0.0} & \score{0.0} & \score{0.0} & \score{0.0} \\
GPT-5 & \score{0} & \score{100} & \score{67} & \score{0} & \score{0} & \score{33.4} & \score{60.1} & \score{0.0} & \score{1.4} \\
\bottomrule
\end{tabular}
\end{table}


\subsection{Processing Speed (S)}

Table~\ref{tab:S_percent} shows unchanged speed profiles across models: strong document-centric throughput (Re/Wr/Num) but near-zero simple/choice reaction and inspection time. GM/WGM (\(\score{0.01}\)) indicate real-time responsiveness remains a bottleneck.

\begin{table}[H]
\footnotesize
\setlength{\tabcolsep}{3.5pt} 
\centering
\caption{Processing Speed (S): Normalized Subdomains with Weights and Aggregates}
\label{tab:S_percent}
\resizebox{\columnwidth}{!}{%
\begin{tabular}{lcccccccccc|cccc}
\toprule
\textbf{Model} & PS-S & PS-C & Re & Wr & Num & SRT & CRT & IT & CS & PF & \textbf{AM} & \textbf{WAM} & \textbf{GM} & \textbf{WGM} \\
\midrule
\textbf{Weight} & 10 & 10 & 10 & 10 & 10 & 10 & 10 & 10 & 10 & 10 & -- & -- & -- & -- \\
\midrule
GPT-4 & \score{0} & \score{0} & \score{100} & \score{100} & \score{100} & \score{0} & \score{0} & \score{0} & \score{0} & \score{0} & \score{30.0} & \score{30.0} & \score{0.01} & \score{0.01} \\
GPT-5 & \score{0} & \score{0} & \score{100} & \score{100} & \score{100} & \score{0} & \score{0} & \score{0} & \score{0} & \score{0} & \score{30.0} & \score{30.0} & \score{0.01} & \score{0.01} \\
\bottomrule
\end{tabular}%
}
\end{table}

\subsection{Discussion}

\textbf{AM} captures domain breadth, \textbf{WAM} preserves the weight structure adopted by \citet{hendrycks2025agidefinition}, while \textbf{GM} and \textbf{WGM} emphasize robustness and balance across subskills.  
Across Tables~\ref{tab:K_percent}--\ref{tab:S_percent}, the geometric aggregates consistently expose residual brittleness in multimodal perception, long-term memory (storage and retrieval), and adaptive reasoning—even where GPT-5’s AM/WAM approach human-level performance.  
This divergence between arithmetic and geometric summaries indicates that apparent gains are concentrated in a subset of faculties (e.g., crystallized knowledge and text-centric skills), with limited cross-modal integration and durable memory.

Methodologically, this paper uses the ten normalized domain scores reported by \citet{hendrycks2025agidefinition} as a shared empirical basis.  
We note, however, that these scores inherit several debatable design choices (e.g., potential conflation across subdomains and relatively low weights for adaptive planning).  
The contrast between AM/WAM and GM/WGM within Tables~\ref{tab:K_percent}--\ref{tab:S_percent} further shows that arithmetic aggregation can \emph{inflate} overall impressions of capability relative to coherence-oriented measures.

Therefore, the $\mathrm{AGI}_{\mathrm{AUC}}$ value of approximately $24\%$ for GPT-5 should be interpreted as an \emph{upper bound} on current general competence.  
As demonstrated across domains in Appendix~\ref{appendix:norm}, even the ten constituent cognitive scores display inherent inflation when converted from weighted arithmetic means to geometric means or unweighted aggregates.  
For instance, in auditory processing, the reported weighted mean of roughly $60\%$ drops to $33\%$ under unweighted averaging and collapses near zero when using geometric means (weighted or unweighted).  
A similar pattern holds for processing speed (from $30\%$ to near zero under geometric averaging), visual processing (from $40\%$ weighted to $29\%$ unweighted and again near zero geometrically), and the working- and long-term memory domains, which remain effectively null regardless of weighting.  
Even on-the-spot reasoning, nominally at $70\%$ by arithmetic mean, falls sharply to $19\%$ (weighted geometric) or $5\%$ (unweighted geometric), revealing fragile compositional performance masked by additive aggregation.  
These discrepancies underscore that the nominal $24\%$ $\mathrm{AGI}_{\mathrm{AUC}}$ should be viewed as conservative yet still optimistic, given the inflation present at the domain-score level.

In the main paper, we therefore focus on the \emph{aggregation principle} rather than revising the underlying measurements.  
The reported domain scores are treated as reasonable, though potentially optimistic, proxies for current model ability, and are used to demonstrate how coherence-aware aggregation (culminating in $\mathrm{AGI}_{\mathrm{AUC}}$) provides a more faithful summary of systemic competence.  
Future work should revisit the domain taxonomy, explore nonlinear scaling and adaptive weightings based on empirical dependencies, and develop diagnostic tools to identify which deficiencies most constrain $\mathrm{AGI}_{\mathrm{AUC}}$, ultimately guiding targeted architectural or training interventions.

\section{Beyond the CHC-domain Scores}
\label{appendix_beyond}

\begin{table*}[t]
\centering
\scriptsize
\begin{tabularx}{\textwidth}{l X c c c c}
\toprule
\textbf{Benchmark} & \textbf{Description} & \textbf{Gemini 3 Pro} & \textbf{Gemini 2.5 Pro} & \textbf{Claude Sonnet 4.5} & \textbf{GPT-5.1} \\
\midrule
Humanity’s Last Exam & Academic reasoning & \textbf{37.5} & 21.6 & 13.7 & 26.5 \\
ARC-AGI-2 & Visual reasoning puzzles & \textbf{31.1} & 4.9 & 13.6 & 17.6 \\
GPQA Diamond & Scientific knowledge & \textbf{91.9} & 86.4 & 83.4 & 88.1 \\
AIME 2025 & Mathematics (no tools) & \textbf{95.0} & 88.0 & 87.0 & 94.0 \\
MathArena Apex & Challenging math contest problems & \textbf{23.4} & 0.5 & 1.6 & 1.0 \\
MMMU-Pro & Multimodal understanding \& reasoning & \textbf{81.0} & 68.0 & 68.0 & 76.0 \\
ScreenSpot-Pro & Screen understanding & \textbf{72.7} & 11.4 & 36.2 & 3.5 \\
CharXiv Reasoning & Information synthesis from charts & \textbf{81.4} & 69.6 & 68.5 & 69.5 \\
Video-MMMU & Video-based knowledge acquisition & \textbf{87.6} & 83.6 & 77.8 & 80.4 \\
Terminal-Bench 2.0 & Agentic terminal coding & \textbf{54.2} & 32.6 & 42.8 & 47.6 \\
SWE-Bench Verified & Agentic coding (single attempt) & 76.2 & 59.6 & \textbf{77.2} & 76.3 \\
t$^2$-bench & Agentic tool use & \textbf{85.4} & 54.9 & 84.7 & 80.2 \\
FACTS Benchmark Suite & Grounding, parametric, MM, search & \textbf{70.5} & 63.4 & 50.4 & 50.8 \\
SimpleQA Verified & Parametric knowledge & \textbf{72.1} & 54.5 & 29.3 & 34.9 \\
MMLU & Multilingual QA & \textbf{91.8} & 89.5 & 89.1 & 91.0 \\
Global PIQA & Commonsense reasoning (100 languages) & \textbf{93.4} & 91.5 & 90.1 & 90.9 \\
MRCR v2 (8-needle) & Long-context (128k avg) & \textbf{77.0} & 58.0 & 47.1 & 61.6 \\
\bottomrule
\end{tabularx}
\caption{Comparison of leading large-language and multimodal models on diverse reasoning, multimodal, agentic, and long-context benchmarks. Scores are taken from the publicly released \emph{Gemini 3 Pro Model Evaluation Report} \citep{deepmind2025gemini3pro}.}
\label{tab:benchmark_comparison}
\end{table*}

\begin{figure}[t]
  \centering
  \includegraphics[width=0.55\linewidth]{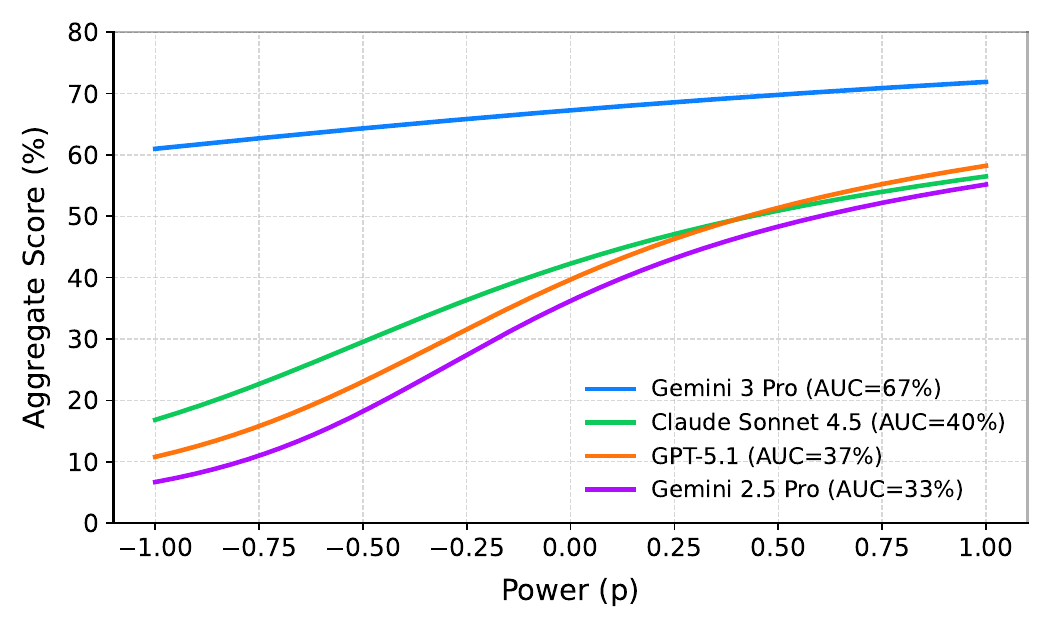}
  \caption{
    Comparison of model performance across aggregation exponents $p$.
    Curves show $\mathrm{AGI}_p$ values derived from the 17 benchmarks in Table~\ref{tab:benchmark_comparison}.
  }
  \label{fig:auc}
\end{figure}

To illustrate that the coherence-based framework extends beyond the CHC taxonomy, we apply the same aggregation methodology to the 17 percentage-based benchmarks reported in Table~\ref{tab:benchmark_comparison}, treating each as an independent capability axis. These benchmarks are drawn from the publicly released \emph{Gemini 3 Pro Model Evaluation Report} \citep{deepmind2025gemini3pro}, which compares Gemini~3~Pro against GPT-5.1, Gemini~2.5~Pro, and Claude Sonnet~4.5 \citep{claude_sonnet4_5}.

These benchmarks span a broad spectrum of reasoning~\citep{phan2025humanity,chollet2025arc,rein2024gpqa,balunovic2025matharena}, multimodal perception~\citep{yue2025mmmu,li2025screenspot,wang2024charxiv,hu2025video}, agentic behavior~\citep{tbench_2025,yang2025swesmith,barres2025tau2}, knowledge and grounding~\citep{haas2025simpleqa,chang2025global}, and long-context understanding.

Let $b_i \in [0,100]$ denote a model’s score on benchmark $i$ for $i=1,\dots,17$. 
We compute $\mathrm{AGI}_p$ using the same continuous family of compensability exponents $p \in [-1,1]$ introduced in Eq.~\eqref{eq:agip}, with stability parameter $\varepsilon = 10^{-6}$. 
We then evaluate the corresponding $\mathrm{AGI}_p$ curves and compute the integrated coherence score $\mathrm{AGI}_{\mathrm{AUC}}$ (Eq.~\ref{eq:agiauc}) using the trapezoidal rule.

\textbf{Results.}  
The 17-benchmark evaluation produces a pattern that closely mirrors the CHC-domain analysis. Positive-$p$ aggregation (including the arithmetic mean at $p=1$) consistently overstates overall capability by masking unevenness across domains, whereas negative-$p$ aggregation sharply penalizes weak dimensions, revealing the true coherence profile of each model.  
Across the $p$-sweep, $\mathrm{AGI}_{\mathrm{AUC}}$ yields a more stable and discriminative ordering than the arithmetic mean: models with similar averages often diverge substantially in coherence, exposing persistent bottlenecks that single-number averages fail to capture.

Notably, Gemini~3~Pro exhibits the highest coherence across the full range of $p$ values. Its $\mathrm{AGI}_p$ curve remains consistently above those of the other models, reflecting improvements not only in overall performances but also in the hardest benchmarks, precisely the regime where coherence penalties matter most. This demonstrates that Gemini~3~Pro’s gains are broad rather than concentrated, leading to a substantially stronger AUC score (67\%).

These findings confirm that the coherence-based evaluation introduced in this work is not dependent on the CHC taxonomy. Instead, it generalizes naturally to heterogeneous benchmark suites, offering a unified and domain-agnostic measure of holistic capability. By rewarding balanced competence and penalizing systemic weaknesses, $\mathrm{AGI}_{\mathrm{AUC}}$ provides a conservative and robust indicator of emerging general intelligence.

It is important to note that any aggregate score, whether derived from 17 benchmarks or hundreds, must be interpreted in light of the benchmark suite it summarizes. No numerical percentage can meaningfully reflect “progress toward AGI’’ unless the underlying tasks collectively span all essential cognitive faculties. When key abilities such as continual learning, long-term memory, or autonomous planning are missing, the resulting aggregate reflects performance only within the tested scope. A system with a bottleneck in any essential faculty cannot be described as achieving a given fraction of AGI, but rather as exhibiting unequal development across capabilities. Within any benchmark suite, however, coherence-based aggregation remains the most informative and principled evaluation method: it highlights imbalances, exposes hidden bottlenecks, and rewards the balance and robustness that are central to general intelligence.

\end{document}